# Knowledge integration for conditional probability assessments

Angelo Gilio
Dip. Metodi e Modelli Matematici
University "La Sapienza"
Via A. Scarpa 10 - 00161 Rome

Fulvio Spezzaferri
Dip. Statistica
University "La Sapienza"
P.le A. Moro 5 - 00185 Rome

## Abstract

In the probabilistic approach to uncertainty management the input knowledge is usually represented by means of some probability distributions. In this paper we assume that the input knowledge is given by two discrete conditional probability distributions, represented by two stochastic matrices P and Q. The consistency of the knowledge base is analyzed. Coherence conditions and explicit formulas for the extension to marginal distributions are obtained in some special cases.

## 1 INTRODUCTION

The uncertainty management in inference systems is usually carried out by representing the partial knowledge through conditional or unconditional probabilities. In this context there are two main problems. The first one is about checking of the consistency of the knowledge base. The second one concerns with the "knowledge integration" process, that is with the extension, in a consistent way, of the initial probabilistic assessment by possibly constructing a complete probabilistic model. A survey of methods used to integrate the input knowledge is presented in Jirousek (1990).

To check consistency we refer to the well known coherence principle of de Finetti (1974), which can be based on the betting scheme or on the penalty criterion. The suitability of this approach has been outlined, for example, in Coletti, Gilio and Scozzafava (1991).

In this paper we consider two random variables X and Y, assuming respectively values $\{x_1,...,x_l\}$, $\{y_1,...,y_m\}$. Denoting by $A_i$ and $B_j$ the events $(X=x_i)$, $(Y=y_j)$, we consider the conditional events $B_j|A_i$, $A_i|B_j$, $i=1,...,l$, $j=1,...,m$. We assume that the knowledge base is represented by two stochastic matrices $P=\{p_{ij}\}$ and $Q=\{q_{ji}\}$, where $p_{ij}=\mathcal{P}(B_j|A_i)$ and $q_{ji}=\mathcal{P}(A_i|B_j)$ are respectively the probabilities of $B_j|A_i$ and $A_i|B_j$. Note that in this paper the concept of conditional event is not used in the same sense as, for example, by Goodman and Nguyen (1988). Following de Finetti's approach, a conditional event A|B is defined as true, false, undetermined, respectively, when A and B are true, A is false and B is true, B is false.

In section 2 some preliminary concepts and results are given. After some general results on coherence, a necessary and sufficient condition for the consistency of the initial assessment (P, Q) is described in section 3. In section 4 we evaluate the probabilities of the marginal events $A_i$ and $B_j$ as a coherent extension of (P, Q). Considering some special cases, necessary and sufficient conditions for the coherence of P and Q and explicit formulas for the marginals are obtained. Our results are based on a valuable relation on conditional probabilities introduced by Császár (1955). A particular version of this relation, named "generalized Bayes theorem", has been used in a recent paper of Amarger, Dubois and Prade (1991) as a starting point in order to obtain the best bracketing of a conditional probability of interest, on the base of the bracketing of other conditional probabilities.

## 2  SOME PRELIMINARIES

In this section we introduce some preliminary results which will be used in what follows.

### 2.1  Generalized atoms

Given a probability assessment $P=(p_1,\ldots,p_n)$ for a family $\mathcal{F}$ of n conditional events $E_1|H_1,\ldots,E_n|H_n$, denote by $C_1,\ldots,C_s$ the atoms generated by $E_i$, $H_i$, $i=1,\ldots,n$, and contained in $H_1\cup\ldots\cup H_n$. The generalized atoms $Q_1,\ldots,Q_s$ relative to $\mathcal{F}$ and P are defined as $Q_h=(\alpha_{h1},\ldots,\alpha_{hn})$, where $\alpha_{hi}=1$ or 0 or $p_i$, according to whether $C_h\subseteq E_iH_i$ or $C_h\subseteq\overline{E}_iH_i$ or $C_h\subseteq\overline{H}_i$, $h=1,\ldots,s$ (Gilio, 1990, 1992).

### 2.2  De Finetti's coherence principle

Denote by $\mathcal{K}$ an arbitrary family of conditional events and by $\mathcal{P}$ a real function defined on $\mathcal{K}$. Given n conditional events $E_1|H_1,\ldots,E_n|H_n$ belonging to $\mathcal{K}$, letting $p_i=\mathcal{P}(E_i|H_i)$, $i=1,\ldots,n$, we associate the following loss function to the point $P=(p_1, p_2, \ldots, p_n)$

$$\mathcal{L} = \sum_{1}^{n} {}_i H_i(E_i-p_i)^2,$$

where the same symbol denotes the event and its indicator. Moreover, let $L_1,\ldots,L_s$ be the values of $\mathcal{L}$ corresponding to the atoms $C_1,\ldots,C_s$. Based on the penalty criterion (see de Finetti, 1974), in Gilio (1990) the following definition is given: the real function $\mathcal{P}$ is a coherent conditional probability if, for every $\mathcal{F}=\{E_1|H_1,\ldots,E_n|H_n\}\subseteq\mathcal{K}$, $n=1,2,\ldots$ for each point $P^*\neq P$ there exists a subscript h such that $L_h^*>L_h$, where $L_1^*,\ldots,L_s^*$ are the values of the loss function corresponding to $P^*$.

### 2.3  Probability extensions

Let $\mathcal{P}$ and $\mathcal{P}^*$ be two conditional probabilities on $\mathcal{K}$ and $\mathcal{K}^*$, where $\mathcal{K}\subseteq\mathcal{K}^*$. Then $\mathcal{P}^*$ is called an extension of $\mathcal{P}$ if, for each $E|H\in\mathcal{K}$, $\mathcal{P}^*(E|H)=\mathcal{P}(E|H)$. Moreover, if $\mathcal{P}$ is coherent there exists (at least) a coherent extension $\mathcal{P}^*$. On the contrary, given $\mathcal{P}^*$ coherent on $\mathcal{K}^*$ its restriction $\mathcal{P}$ on $\mathcal{K}$ is coherent too.

### 2.4  Császár's condition

Given a conditional probability $\mathcal{P}$, defined on $\mathcal{E}\times\mathcal{H}$ where $\mathcal{E}$ is an algebra of events and $\mathcal{H}\subset\mathcal{E}$ is a non empty family of events not containing $\emptyset$, the family $\mathcal{H}$ is defined additive if, for each $H_1, H_2\in\mathcal{H}$ it is $H_1\cup H_2\in\mathcal{H}$. Moreover, $\mathcal{H}$ is defined $\mathcal{P}$-quasi additive (Császár, 1955) if, for each $H_1, H_2\in\mathcal{H}$, there exists $K\in\mathcal{H}$, $H_1\cup H_2\subseteq K$, such that
$$\mathcal{P}(H_1|K) + \mathcal{P}(H_2|K) > 0 \ .$$
In the same paper (Theorem 8.5) Császár has also shown the equivalence of the propositions:
i) the following condition is satisfied:

$$(1) \qquad \prod_{i=1}^{n}\mathcal{P}(E_i|H_i) = \prod_{i=1}^{n}\mathcal{P}(E_i|H_{i+1}),$$

where $E_i\in\mathcal{E}$, $H_i\in\mathcal{H}$, $E_i\subseteq H_iH_{i+1}$, and $H_{n+1}=H_1$;
ii) there exists an extension of $\mathcal{P}$ to $\mathcal{P}^*$ defined on $\mathcal{E}\times\mathcal{H}^*$, where $\mathcal{H}^*$ is an additive class containing $\mathcal{H}$;
iii) there exists an extension of $\mathcal{P}$ to $\mathcal{P}^*$ defined on $\mathcal{E}\times\mathcal{H}^*$, where $\mathcal{H}^*$ is a $\mathcal{P}$-quasi additive class containing $\mathcal{H}$.

We observe that, when $E_i=H_iH_{i+1}$ and all the involved probabilities are positive, Császár's condition (1) reduces to the "generalized Bayes theorem", introduced in Amarger, Dubois and Prade (1991).

### 2.5  Connected matrices

In a paper of Spezzaferri (1981) the following definition is given: an $l\times m$ matrix P is connected if there are no row and column permutations which change P to the form

$$\begin{bmatrix} A_{r,s} & 0_{r,m-s} \\ 0_{l-r,s} & B_{l-r,m-s} \end{bmatrix},$$

for some r, s, $1\leq r<l$, $1\leq s<m$.

## 3  COHERENCE OF (P, Q)

Assuming positive probabilities for the conditioning events, the consistency checking of the initial assessment can be carried out using linear programming. More in general, without assuming strictly positive probabilities for the conditioning events, the consistency can be checked using the following result (Gilio, 1990, 1992).

> Proposition 1
> Given an assessment $p_1, p_2,\ldots, p_n$ for n conditional events $E_1|H_1, E_2|H_2,\ldots, E_n|H_n$,

a necessary and sufficient condition for its coherence is that, for each subset $I \subseteq \{1,2,...,n\}$, the point $P_I = (p_i, i \in I)$ is a convex linear combination of the generalized atoms relative to the family $\mathcal{F}_I = \{E_i | H_i, i \in I\}$ and to the point $P_I$.

The family of events considered in Proposition 1 is not required to have any algebraic property. An alternative way to check coherence when the family of events involved has a particular structure is stated in the following theorem (Rigo, 1988, Berti, Regazzini and Rigo, 1990).

**Theorem 2**
A conditional probability $\mathcal{P}$, defined on $\mathcal{E} \times \mathcal{H}$ where $\mathcal{E}$ is an algebra of events and $\mathcal{H} \subset \mathcal{E}$ is a non empty family of events not containing $\emptyset$, is coherent if and only if, for each n, condition (1) is satisfied.

A different proof of Theorem 2, based on the generalized atoms (see section 2.1), is given in Gilio and Spezzaferri (1992). In the same paper, from the previous theorem, the following result on coherence of $(P, Q)$ is obtained.

**Corollary 3.**
The conditional probability assessment $(P, Q)$, where P and Q are stochastic matrices, is coherent if and only if, for each $n \leq \min(l, m)$

(2) $p_{i_1 j_1} q_{j_1 i_2} \cdots p_{i_h j_h} q_{j_h i_{h+1}} \cdots p_{i_n j_n} q_{j_n i_1} =$

$p_{i_1 j_n} q_{j_n i_n} \cdots p_{i_{h+1} j_h} q_{j_h i_h} \cdots p_{i_2 j_1} q_{j_1 i_1}$,

where $i_k \in \{1,2,...,l\}$, $j_k \in \{1,2,...,m\}$, $k=1,2,...,n$, and $i_k \neq i_h$, $j_k \neq j_h$ for $k \neq h$.

The proof of Corollary 3 follows from Theorem 2 extending the assessment $(P, Q)$ to a conditional probability $\mathcal{P}$ on $\mathcal{E} \times \mathcal{H}$, where $\mathcal{E}$ is the algebra generated by the family $\mathcal{H} = \{A_i, B_j, i=1,...,l, j=1,...,m\}$ and observing that, in this case, condition (1) reduces to (2).

## 4  EXTENSION TO MARGINAL DISTRIBUTIONS

In many applications, in order to integrate the knowledge, it can be required to compute the probabilities of the events $A_i$ and $B_j$, that is the marginal distributions of the random variables X and Y. If the assessment $(P, Q)$ is coherent, then $(P, Q)$ can be extended to a conditional probability $\mathcal{P}$ on $\mathcal{E} \times \mathcal{H}$ satisfying condition (2), where $\mathcal{E}$ is the algebra generated by the family $\mathcal{H} = \{A_i, B_j, i=1,...,l, j=1,...,m\}$. Now, from the quoted Császár's Theorem 8.5, $\mathcal{P}$ can be extended to $\mathcal{P}^*$ defined on $\mathcal{E} \times \mathcal{H}^*$, where $\mathcal{H}^*$ is an additive class containing $\mathcal{H}$ and the certain event $\Omega = A_1 \cup ... \cup A_l = B_1 \cup ... \cup B_l$. Therefore, there exists at least an extension of $\mathcal{P}$ to the events $A_i | \Omega$, $B_j | \Omega$, $i=1,...,l$, $j=1,...,m$, providing the required marginal distributions of X and Y. Denoting this extension by the same symbol $\mathcal{P}$ and defining $f_i = \mathcal{P}(A_i|\Omega)$ and $g_j = \mathcal{P}(B_j|\Omega)$, $i=1,...,l$, $j=1,...,m$, we observe that the marginal distributions $\{f_i\}$, $\{g_j\}$ satisfy the following linear system:

(3)
$$p_{ij} f_i = q_{ji} g_j, \quad \sum f_i = \sum g_j = 1,$$
$$f_i \geq 0, \ g_j \geq 0, \ i=1,...,l, \ j=1,...,m.$$

Now, the checking of coherence of P and Q will be considered in some special cases and the problem of the uniqueness and computation of marginal distributions will be examined. To this regard, it will be proved useful to examine preliminarly the compatibility of (3), as a necessary condition for coherence of $(P, Q)$.

### 4.1  P AND Q WITH TWO POSITIVE COLUMNS

Assume that P and Q satisfy the following property:

**Property 4**
There exist two indices $h_o$, $k_o$ such that $p_{ih_o} > 0$, $q_{jk_o} > 0$, $i=1,...,l$, $j=1,...,m$.

Then, we have the following theorem (Gilio and Spezzaferri, 1992).

**Theorem 5**
If P and Q satisfy Property 4, the system (3) is compatible if and only if the following condition is satisfied:

(4) $p_{ij} q_{jk_o} p_{k_o h_o} q_{h_o i} = p_{ih_o} q_{h_o k_o} p_{k_o j} q_{ji}$,

$i=1,2,...,l$, $j=1,2,...,m$.

Moreover, if (3) is compatible the unique solution is given by

(5) $f_i^* = (q_{h_o i}/p_{ih_o}) \left( \sum_r q_{h_o r}/p_{r h_o} \right)^{-1}$,

$$g_j^* = (p_{k_o j}/q_{jk_o})\left(\sum_r p_{k_o r}/q_{rk_o}\right)^{-1},$$

$$i=1,\ldots,l,\ j=1,\ldots,m.$$

We observe that the compatibility of system (3) does not imply that condition (2) is satisfied. Therefore the coherence of (P, Q) is not assured, as it is shown in the following example. Let

$$P = \begin{bmatrix} 1 & 0 & 0 \\ 1/3 & 1/3 & 1/3 \\ 1/3 & 1/3 & 1/3 \end{bmatrix} \quad Q = \begin{bmatrix} 1 & 0 & 0 \\ 1/4 & 1/2 & 1/4 \\ 1/4 & 1/4 & 1/2 \end{bmatrix}.$$

Property 4 is satisfied, with $h_o = k_o = 1$, and the unique solution is: $f_i = g_i = 1\ (0)$ if $i=1\ (>1)$. However condition (2) does not hold for $n=2$, $i_1 = j_1 = 2$, $i_n = j_n = 3$, so that the assessment (P, Q) is not coherent.

In order to achieve coherence a result of Berti, Regazzini and Rigo (1991, Corollary 1.3.) can be used in the framework of Theorem 2. It states that if $\mathcal{H} = \prod_1 \cup \{\Omega\} \cup \prod_2$, where $\prod_1$ and $\prod_2$ are two partitions of $\Omega$ contained in $\mathcal{E}$, $\mathcal{P}$ is coherent if condition (2) is satisfied for the $H_i$'s such that $\mathcal{P}(H_i)=0$.

This result can be applied in our case defining $\prod_1 = \{A_i,\ i=1,\ldots,l\}$, $\prod_2 = \{B_j,\ j=1,\ldots,m\}$ and checking (2) for $i_k$, $j_k$ such that $f_{i_k}^* = g_{j_k}^* = 0$. Note that $f_i^*\ (g_j^*) > 0$ only if $q_{h_o i}\ (p_{k_o j}) > 0$.

Finally, in order to check the coherence of the assessment (P, Q) we have to verify condition (4), which amounts to compatibility of (3), and then condition (2) for $i_k$, $j_k$ such that $q_{h_o i_k} = p_{k_o j_k} = 0$.

### 4.2  P AND Q STRICTLY POSITIVE

The important case when P and Q are strictly positive is now considered. In this case the compatibility of system (3) does imply the coherence of (P, Q). In fact, if the system (3) is compatible, its solution is strictly positive. Therefore the checking of coherence of (P, Q) reduces to verify condition (4).

The following alternative solution of system (3), equivalent to (5), has been obtained in Spezzaferri (1981)

$$f_i^* = \left(\sum_h p_{ih}/q_{hi}\right)^{-1},\ g_j^* = \left(\sum_k q_{jk}/p_{kj}\right)^{-1},$$

$$i=1,\ldots,l,\ j=1,\ldots,m.$$

### 4.3  P AND Q SUCH THAT $p_{ij} > 0$ IF AND ONLY IF $q_{ji} > 0$

Assume that P and Q satisfy the following property:

**Property 6**
For any pair of subscripts (i, j) it is $p_{ij} > 0$ if and only if $q_{ji} > 0$.

In the paper of Spezzaferri (1981, Theorem 3) it is shown that if the system (3) is compatible, then the solution is unique if and only if P (and therefore Q) is connected.

#### 4.3.1  P AND Q CONNECTED

Assume that P and Q are connected. Let $U = \{u_{ij}\}$ be an $l \times m$ non negative connected matrix with exactly $l+m-1$ positive elements, each of which coincides with the corresponding element of P. Let $V = \{v_{ji}\}$ be an $m \times l$ non negative connected matrix obtained in a similar way from Q.

Consider the matrices $T = (UV)^{\gamma-1}U$, $Z = (VU)^{\gamma-1}V$, where $\gamma$ is the smallest integer such that the above matrices are strictly positive. Then, in the quoted paper of Spezzaferri (Theorem 4) it is proved that $\gamma \leq \min(l, m)$ and

(i) the system (3) is compatible if and only if the following condition is satisfied:

(6) $\quad t_{ij}/z_{ji} = p_{ij}/q_{ji},\ \forall\ (i,j): p_{ij} > 0;$

(ii) the (unique and positive) solution is

$$f_i^* = \left(\sum_h t_{ih}/z_{hi}\right)^{-1},\ g_j^* = \left(\sum_k z_{jk}/t_{kj}\right)^{-1},$$

where $t_{ij}$ and $z_{ji}$ are the elements of T and Z.

As in section 4.2, if the system (3) is compatible, its solution is strictly positive. Therefore the compatibility of system (3) implies the coherence of (P, Q) and the checking of coherence reduces to verify condition (6).

Finally, in order to check the coherence of the assessment (P, Q) we have to define two connected matrices U and V, to compute T and Z and to verify condition (6).

Note that in all the above cases, if the assessment (P, Q) is coherent its extension to the marginal distributions is unique. On the contrary, in the case examined in the next section this property does not hold.

### 4.3.2 P AND Q NOT CONNECTED

Assume that P and Q are not connected. Therefore we have

$$P = \begin{bmatrix} A_{r,s} & 0_{r,m-s} \\ 0_{l-r,s} & B_{l-r,m-s} \end{bmatrix},$$

$$Q = \begin{bmatrix} C_{s,r} & 0_{s,l-r} \\ 0_{m-s,r} & D_{m-s,l-r} \end{bmatrix},$$

for some r, s, $1 \leq r < l$, $1 \leq s < m$, where A, B, C, D are stochastic matrices.

We have the following result:

**Proposition 7**
The assessment (P, Q) is coherent if and only if the two assessments (A, C) and (B, D) are coherent.

Proof. Observe that condition (2) of Corollary (3), applied to the actual (P, Q), reduces to the identity 0=0 if among the subscripts $i_k$ or $j_k$, k=1,2,...,n, there is a pair $(i_{k_1}, i_{k_2})$ or $(j_{k_1}, j_{k_2})$ such that

$$1 \leq i_{k_1} \leq r \text{ and } r+1 \leq i_{k_2} \leq l$$
or
$$1 \leq j_{k_1} \leq s \text{ and } s+1 \leq j_{k_2} \leq m.$$

Therefore condition (2) applied to (P, Q) is equivalent to the same condition applied separately to (A, C) and (B, D).

Let us now assume that A, B (and therefore C, D) are connected. Then, we can apply the procedure described in section 4.3.1 to the assessments (A, C) and (B, D). Consider the systems

(7) $\quad p_{ij}f_i = q_{ji}g_j, \quad \sum f_i = \sum g_j = 1,$
$\quad f_i \geq 0, \ g_j \geq 0, \ i=1,...,r, \ j=1,...,s;$

(8) $\quad p_{ij}f_i = q_{ji}g_j, \quad \sum f_i = \sum g_j = 1,$
$\quad f_i \geq 0, \ g_j \geq 0, \ i=r+1,...,l, \ j=s+1,...,m.$

If (7) and (8) are compatible (i.e., the corresponding matrices satisfy condition (6)), then (A, C) and (B, D) are both coherent and, by Proposition 7, (P, Q) is coherent too. Moreover, denoting respectively by $\bar{f}_1, ..., \bar{f}_r, \bar{g}_1, ..., \bar{g}_s$ and by $\bar{f}_{r+1}, ..., \bar{f}_l, \bar{g}_{s+1}, ..., \bar{g}_m$ the solutions of (7) and (8), for each $\theta \in (0,1)$, the set of values

(9) $\quad f_i^* = \theta \bar{f}_i, \ i=1,...,r, \ f_i^* = (1-\theta)\bar{f}_i, \ i=r+1,...,l,$
$\quad g_j^* = \theta \bar{g}_j, \ j=1,...,s, \ g_j^* = (1-\theta)\bar{g}_j, \ j=s+1,...,m,$

is a positive solution of system (3), providing (as in sections 4.2 and 4.3.1) a coherent extension to the marginals of the assessment (P, Q). Note that, from the closure property of coherence with respect to the limit, the two sets of values (9) corresponding to $\theta=0$, $\theta=1$ provide coherent marginals too.

To remark that in general compatibility is not equivalent to coherence, we observe that if only one of the two systems (7), (8) is compatible, e.g. (7), then the assessment (P, Q) is not coherent. However the system (3) admits the solution (9) with $\theta=1$.

Finally, if A or B (and therefore C or D) is not connected we have to apply to (A, C) or (B, D) the same procedure developed in this section to analyze (P, Q). This process can be iterated to obtain a partition of P and Q in connected matrices.

### 5 CONCLUSIONS

In this paper it has been assumed that the knowledge base consists of two discrete conditional distributions, represented by two stochastic matrices P and Q. The problems of coherence of the initial assessment and that of its extension to the marginal distributions have been examined. To investigate coherence a key role has been played by Császár's condition. It has been shown that to study the previous problems it is useful to examine the compatibility of a suitable system associated to P and Q. In some special cases necessary and sufficient conditions for the coherence of P and Q and explicit formulas for the marginals have been given. To analyze the relationship between compatibility and coherence a useful tool has been the concept of connected matrix.

## REFERENCES

S. Amarger, D. Dubois and H. Prade (1991), Constraint propagation with imprecise conditional probabilities, Proc. of The 7th Conf. on Uncertainty in Artificial Intelligence, Eds. B.D. D'Ambrosio, P. Smets, P.P. Bonissone, Morgan Kaufmann Publ., San Mateo, California, 26-34.

P. Berti, E. Regazzini and P. Rigo (1990), de Finetti's coherence and complete predictive inferences, Working paper, IAMI 90.5.

P. Berti, E. Regazzini and P. Rigo (1991), Coherent statistical inference and Bayes theorem, Annals of Statistics, Vol. 19, N. 1, 366-381.

G. Coletti, A. Gilio and R. Scozzafava (1991), Conditional events with vague information in expert systems, Lecture Notes in Computer Science, 521, B. Bouchon-Meunier, R. R. Yager, L. A. Zadeh (Eds.), Springer Verlag, 106-114.

Á. Császár (1955), Sur la structure des espaces de probabilité conditionnelle, Acta Mathematica Academiae Scientiarum Hungaricae, 46, 337-361.

B. de Finetti (1974). Theory of probability, John Wiley, New York.

A. Gilio (1990), Criterio di penalizzazione e condizioni di coerenza nella valutazione soggettiva della probabilita', Boll. Un. Mat. Ital., vol. 4-B, n. 3, Serie 7, 645-660.

A. Gilio (1992), $C_o$ - coherence and extensions of conditional probabilities, Bayesian Statistics 4, J. M. Bernardo, J. O. Berger, A. P. Dawid and A. F. M. Smith (Eds.), Oxford University Press, 1992. In press.

A. Gilio and F. Spezzaferri (1992), Coherence and extensions of conditional probability assessments. Working paper.

I. R. Goodman and H. T. Nguyen (1988), Conditional objects and modelling of uncertainties, Fuzzy computing - Theory, Hardware and Applications, M. M. Gupta, T. Yamakawa (Eds.), North-Holland, 119-138.

R. Jirousek (1990), A survey of methods used in probabilistic expert systems for knowledge integration, AI 1989 Conference, Prague, Butterworth & Co. (Publ.) Ltd, 7-12.

P. Rigo (1988), Un teorema di estensione per probabilita' condizionate finitamente additive, Atti della XXXIV Riunione Scientifica della S.I.S., Siena, Vol. 2(1), 27-34.

F. Spezzaferri (1981), Reconstructing marginals from conditional distributions, Metron, Vol. XXXIX, N. 3-4, 161-176.